\documentclass[letterpaper, 10 pt, conference]{ieeeconf}  % Comment this line out if you need a4paper

\IEEEoverridecommandlockouts                              % This command is only needed if 
\usepackage{graphics} % for pdf, bitmapped graphics files
\usepackage{epsfig} % for postscript graphics files
\usepackage{mathptmx} % assumes new font selection scheme installed
\usepackage{times} % assumes new font selection scheme installed
\usepackage{amsmath} % assumes amsmath package installed
\usepackage{amssymb}  % assumes amsmath package installed
\usepackage{subfigure}
\usepackage{multirow}
\usepackage{subcaption}
\usepackage{float}
\usepackage{ctable}
\usepackage{cuted}
\usepackage{colortbl}
\usepackage{algorithm}
\usepackage{algpseudocode}
\usepackage{titlesec}
\usepackage{soul}

\usepackage[numbers,sort&compress]{natbib}
\usepackage{eucal} % Load this in your preamble
\usepackage[colorlinks,linkcolor=blue]{hyperref}

\usepackage{etoolbox}
\AtBeginEnvironment{thebibliography}{\footnotesize}

\newcommand{\lixin}[1]{#1}

\title{\LARGE \bf
TrajBooster: Boosting Humanoid Whole-Body Manipulation via Trajectory-Centric Learning
}

\author{\begin{tabular}{c}
    Jiacheng Liu\textsuperscript{1,2,4*}, Pengxiang Ding\textsuperscript{1,2*}, Qihang Zhou\textsuperscript{3,4}, Yuxuan Wu\textsuperscript{3,4}, Da Huang\textsuperscript{3,4}, Zimian Peng\textsuperscript{1,4},  \\
    \newline
    Wei Xiao\textsuperscript{2}, Weinan Zhang\textsuperscript{3,4}, Lixin Yang\textsuperscript{3,4}$^{\dag}$, Cewu Lu\textsuperscript{3,4}$^{\dag}$, Donglin Wang\textsuperscript{2,4}$^{\dag}$ \\
  \end{tabular}
  \thanks{$^{*}$ Equal contribution. $^{\dag}$ Equal advising.}
\thanks{$^{1}$Zhejiang University, $^{2}$Westlake University, $^{3}$Shanghai Jiao Tong University, $^{4}$Shanghai Innovation Institute.}% <-this % stops a space
}

\begin{document}

\maketitle
\thispagestyle{empty}
\pagestyle{empty}

\begin{strip}
\vspace{-21mm}
\centering
\includegraphics[width=\textwidth]{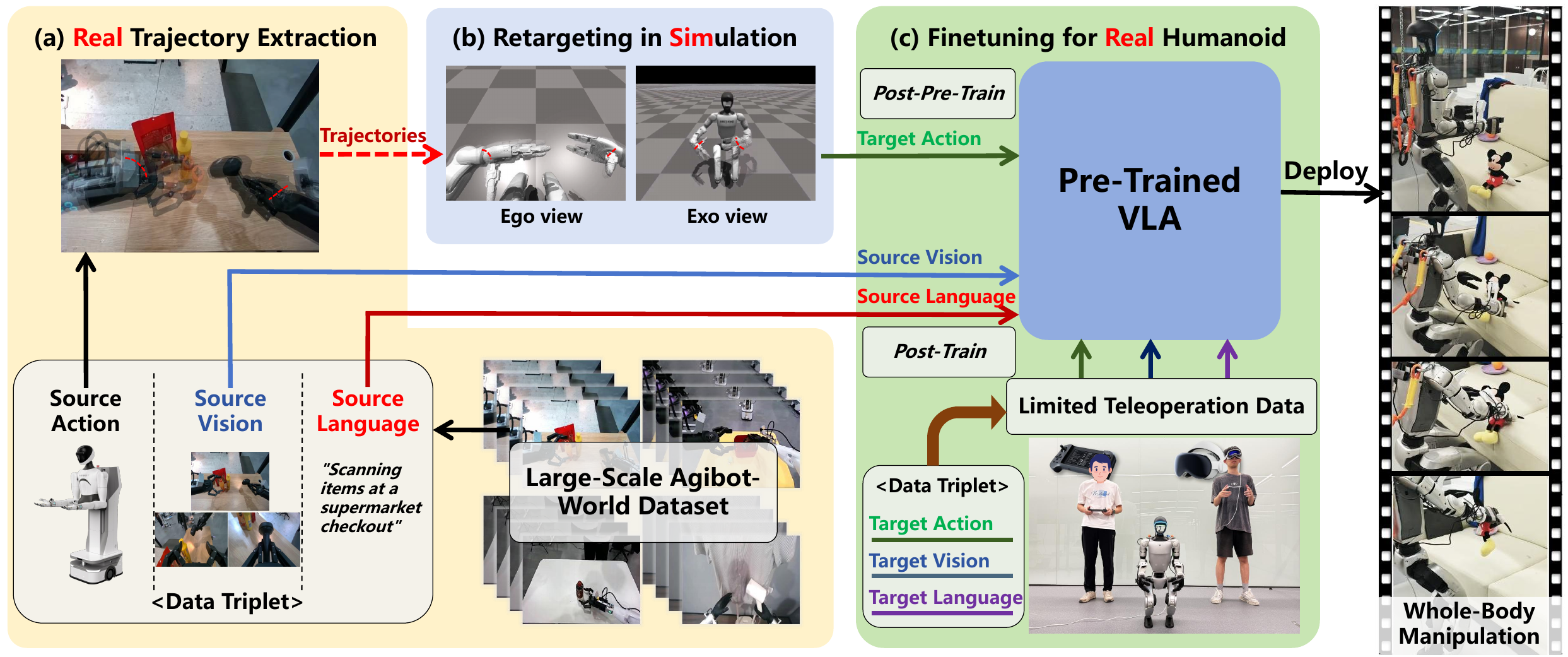}
\captionof{figure}{\textbf{Overview of framework.} 
Our proposed TrajBooster uses abundant existing robot manipulation datasets. It retargets end-effector trajectories from diverse robots to target humanoids via a retargeting model. We then \textit{post-pre-train} a pre-trained VLA with such \textbf{large-scale retargeted data} before \textit{final post-training} with \textbf{minimal real-world data}. This approach reduces the burden of human teleoperation while improving action space comprehension and  zero-shot skill transfer capability. }
\label{fig:TrajBooster_overview}
\vspace{-10px}
\end{strip}

%%%%%%%%%%%%%%%%%%%%%%%%%%%%%%%%%%%%%%%%%%%%%%%%%%%%%%%%%%%%%%%%%%%%%%%%%%%%%%%%
\begin{abstract}
Recent Vision-Language-Action (VLA) models show potential to generalize across embodiments but struggle to quickly align with a new robot’s action space when high-quality demonstrations are scarce, especially for bipedal humanoids. We present TrajBooster, a cross-embodiment framework that leverages abundant wheeled-humanoid data to boost bipedal VLA. Our key idea is to use end-effector trajectories as a morphology-agnostic interface. TrajBooster (i) extracts 6D dual-arm end-effector trajectories from real-world wheeled humanoids, (ii) retargets them in simulation to Unitree G1 with a whole-body controller trained via a heuristic-enhanced harmonized online DAgger to lift low-dimensional trajectory references into feasible high-dimensional whole-body actions, and (iii) forms heterogeneous triplets that couple source vision/language with target humanoid-compatible actions to post-pre-train a VLA, followed by only 10 minutes of teleoperation data collection on the target humanoid domain. Deployed on Unitree G1, our policy achieves beyond-tabletop household tasks, enabling squatting, cross-height manipulation, and coordinated whole-body motion with markedly improved robustness and generalization. Results show that TrajBooster allows existing wheeled-humanoid data to efficiently strengthen bipedal humanoid VLA performance, reducing reliance on costly same-embodiment data while enhancing action space understanding and zero-shot skill transfer capabilities. 
For more details, please refer to our webpage \href{https://jiachengliu3.github.io/TrajBooster/}{https://jiachengliu3.github.io/TrajBooster}.

\end{abstract}

%%%%%%%%%%%%%%%%%%%%%%%%%%%%%%%%%%%%%%%%%%%%%%%%%%%%%%%%%%%%%%%%%%%%%%%%%%%%%%%%
\section{INTRODUCTION}
Recent advances have markedly advanced humanoid manipulation~\cite{yuan2025being,yang2025egovla,bjorck2025gr00t,qiu2025humanoid,bu2025agibot}. Building on this progress, Vision-Language-Action (VLA) models equip humanoid robots to autonomously perform a broad range of household tasks with improved reliability and generalization.

\begin{figure}[t!] 
\centering
\begin{minipage}{0.24\textwidth}
  \centering
  \includegraphics[width=\linewidth]{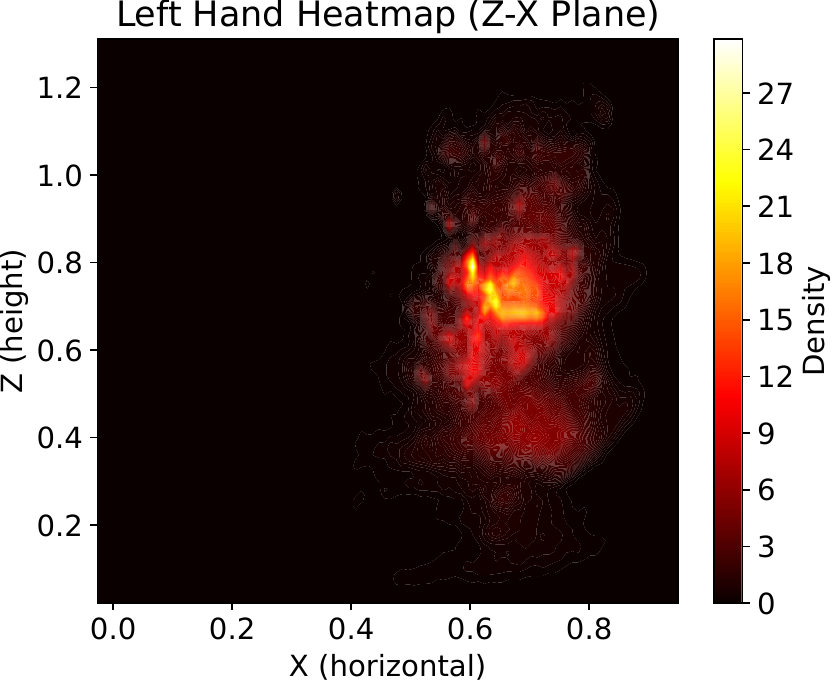}
\end{minipage}
\hfill
\begin{minipage}{0.24\textwidth}
  \centering
  \includegraphics[width=\linewidth]{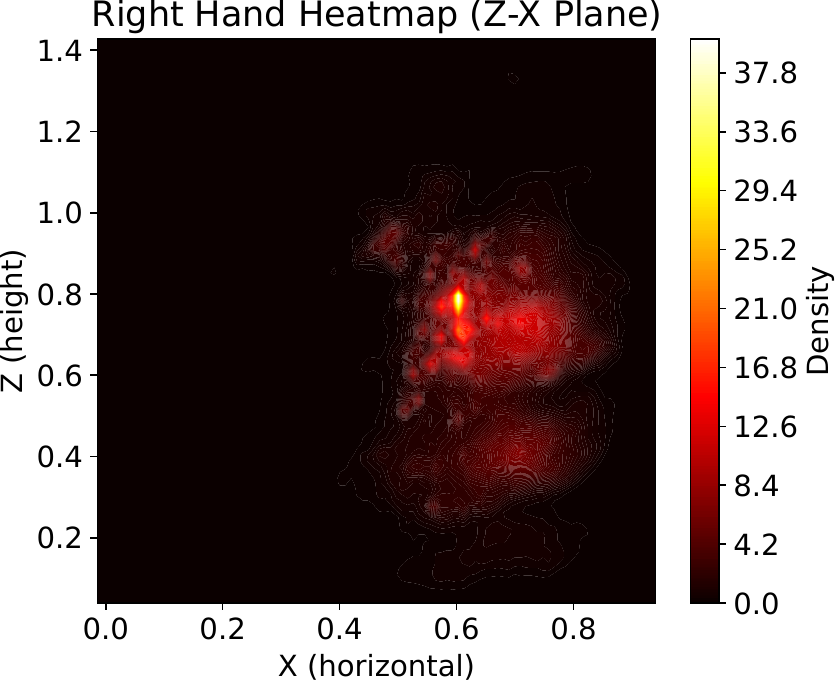} 
\end{minipage}
\caption{\textbf{Hand position heatmaps in Z-X plane from Agibot-World Beta dataset.} Visualization of left and right hand distributions using \textbf{Kernel Density Estimation} (KDE). The X-axis positive direction aligns with the robot's forward heading, while the Z-axis positive direction opposes gravity (upward).}
\label{fig:hot-map}
\vspace{-15px}
\end{figure}

Among them, wheeled humanoid robots have particularly excelled at household tasks that demand coordinated whole-body movements—such as squatting and reaching across varying heights—highlighting the practical reach and dexterity required in real homes. Evidence from the Agibot-World Beta dataset~\cite{bu2025agibot} shows end-effector trajectories concentrated between 0.2–1.2 m (Fig.~\ref{fig:hot-map}), underscoring that everyday domestic tasks demand versatile manipulation across an extended workspace, well beyond tabletop.
By contrast, bipedal humanoids must manipulate with the upper body while maintaining dynamic balance with the lower body, making this wide-range whole-body manipulation particularly challenging. 

Meanwhile, prior VLA research has largely focused on locomotion in complex environments~\cite{ding2025humanoid,xue2025leverb} or on tabletop manipulation~\cite{yang2025egovla,bjorck2025gr00t}, leaving a critical gap: enabling wide-range, whole-body manipulation for bipedal humanoids.

Achieving this capability requires large-scale demonstrations, yet data collection remains the bottleneck. Existing teleoperation pipelines require expensive infrastructure and expert operators, and typically yield datasets that are small, and limited in diversity across different scenes and tasks. As a result, VLAs struggle to align, during post-training, with the action spaces of new humanoid platforms. While pretraining on heterogeneous robot corpora helps, it cannot replace high-quality, humanoid-relevant, whole-body demonstrations with sufficient coverage. Consequently, current systems remain inadequate for wide-range manipulation.

We address this problem with \textbf{TrajBooster}, a cross-embodiment framework (Fig.~\ref{fig:TrajBooster_overview}) that leverages the morphology-agnostic nature of end-effector \textbf{trajectories} to transfer demonstrations from wheeled to bipedal humanoids, mitigating data scarcity in bipedal VLA fine-tuning and thus \textbf{boosting} VLA action space comprehension and task generalization for whole-body manipulation on our target bipedal humanoid.   Our key insight is that, despite morphological differences, end-effector trajectories provide a shared interface that can bridge \lixin{the joint-space gap between} embodiments. Using large-scale data from the wheeled humanoid Agibot G1, we indirectly enhance VLA training for the bipedal Unitree G1 via a real-to-sim-to-real pipeline.
\lixin{Our TrajBooster framework consists of three steps:}

\textbf{1) Real Trajectory Extraction:} the process begins with extracting end-effector trajectories from the source robots. 
Rather than directly mapping full-body motions to the target humanoid, TrajBooster utilizes the 6D coordinates of dual-arm end-effectors as the goal, enabling a retargeting model within the Isaac Gym simulator to achieve whole-body motion retargeting by tracking this goal.

\textbf{2)} \textbf{Retargeting in Simulation:} the retargeting model is trained with our heuristic-enhanced harmonized online DAgger algorithm for the target humanoid Unitree G1 to track these reference trajectories using whole-body control. Through this process, the humanoid learns to coordinate its joints such that its end-effectors follow the retargeted goals, effectively mapping low-dimensional reference signals into feasible whole-body high-dimensional actions. This stage generates a large volume of action data that is compatible with the morphology of the real-world target humanoid.

\textbf{3) Finetuning for real humanoid:} using this newly generated data, TrajBooster constructs heterogeneous triplets in the form of $\langle$\textbf{source} vision, \textbf{source} language, \textbf{target} action$\rangle$, which link perceptual inputs with humanoid-compatible behaviors. The resulting synthetic dataset is then used to post-pre-train an existing VLA model. \lixin{Subsequently,} with just 10 minutes of additional real-world teleoperation data, \textit{i.e} $\langle$\textbf{target} vision, \textbf{target} language, \textbf{target} action$\rangle$, the post-pre-trained VLA is fine-tuned and deployed on the Unitree G1 across a wide spectrum of whole-body manipulation tasks.

To sum up, our contributions are three folds:
\begin{itemize}

\item To the best of our knowledge, this is the \textbf{first} work to leverage extensive retargeted action data for fine-tuning and achieve bipedal \textbf{humanoid whole-body manipulation} with a VLA model in the real world.

\item We introduce \textbf{TrajBooster}, \textbf{a real-to-sim-to-real cross-embodiment framework} that converts abundant wheeled-humanoid demonstrations into effective bipedal humanoid training data, using end-effector trajectories as morphology-agnostic signals. This approach enables VLA adaptation with only limited target-domain data, thus mitigating data scarcity for bipedal humanoids.

\item Deployed on Unitree G1 with only \textbf{10 minutes of teleoperation data collection}, we achieve beyond-tabletop household tasks including squatting, cross-height manipulation, and coordinated whole-body motion with markedly improved robustness and generalization, demonstrating that abundant wheeled-humanoid data can efficiently strengthen bipedal VLA performance while reducing reliance on costly same-embodiment data and enhancing zero-shot skill transfer capabilities.

\end{itemize}

\section{RELATED WORKS}
\subsection{Humanoid Whole-body Control}

Recently, research on real-world humanoid whole-body control has made considerable progress, with many works~\cite{ze2025twist, li2025clone, he2024learning, he2024omnih2o, cheng2024expressive, xue2025unified, lu2025mobile, zhang2025unleashing} advancing the field primarily through teleoperation-based approaches. 
A number of studies such as Humanoid-VLA~\cite{ding2025humanoid} and Leverb~\cite{xue2025leverb} have explored autonomous strategies by employing VLA models to generate full-body motions. However, these efforts have mainly focused on coarse-grained control, such as sitting down, waving hand or walking.

In contrast, research on humanoid manipulation tasks has explored action generation via visuomotor policy~\cite{qiu2025humanoid, ze2024generalizable} or VLA model~\cite{yang2025egovla, bjorck2025gr00t}, but these have largely been confined to tabletop scenarios. Such a setting underutilizes the locomotor capabilities of humanoid lower limbs, thereby restricting the operational space of the robot. 
While Homie~\cite{ben2025homie} represents a notable advancement in addressing this limitation through its visuomotor control policy, its practical applicability remains constrained by the requirement to train individual policies for each task, thereby limiting scalability across diverse task scenarios. To overcome this limitation, our approach leverages \textbf{a unified VLA model} to enable real-world \textbf{whole-body manipulation} across multiple tasks on a bipedal humanoid robot, demonstrating versatile manipulation capabilities across a broad spectrum of operational heights.

\subsection{Cross-embodiment Learning}

Cross-embodiment learning seeks to transfer knowledge across agents with heterogeneous morphologies.
Several methods mitigate perceptual discrepancies using inpainting, segmentation, or physics-based rendering~\cite{li2024ag2manip,kareer2024egomimic,li2025h2r,lepert2025masquerade}, effectively aligning observations but remaining limited to the perception level.
Beyond perception, researchers explore embodiment-invariant action abstractions.
Latent action representations~\cite{ye2024latent,chen2024igor,bu2025agibot,chen2025villa} provide coarse-grained, implicit encodings, whereas trajectory-based methods~\cite{qin2022dexmv,yang2025egovla,bi2025h} extract manipulation skills into explicit forms. For instance, DexMV~\cite{qin2022dexmv} maps human 3D hand poses to robot trajectories. While effective, these approaches mainly address dexterous hand–object interactions and do not scale to full-body motion transfer.
Recent work such as~\cite{ha2024umi} generates full-body actions; however, its applicability is constrained by the limitations of quadruped workspace configurations. 

This work addresses the aforementioned limitation and represents the first application to humanoid scenarios, leveraging \textbf{actuator-space 6D pose remapping} across diverse dual-arm robot demonstrations to facilitate cross-embodiment transfer to a target humanoid robot, thereby achieving cross-embodiment bipedal humanoid manipulation with extensive whole-body workspace coverage.

\section{PROPOSED METHOD}
Our proposed TrajBooster,a real-to-sim-to-real pipeline, is illustrated in Fig.~\ref{fig:TrajBooster_overview}. In this section, we 1) first describe the extraction of real trajectories from existing datasets for retargeting. We 2) then introduce the retargeting model architecture and policy learning algorithm. Finally, we 3) detail the adaptation of a pre-trained VLA through a two-step post-training procedure: i) post-pre-training using the retargeted whole-body manipulation data collected in simulation, and ii) post-training with a small amount of teleoperated data collected in the real world.

\subsection{Real Trajectory Extraction}

\label{subsec:data_preprocess}
We utilize manipulation data from Agibot-World beta dataset~\cite{bu2025agibot} as the real-robot data source. This dataset comprises over one million real-robot trajectories, including multi-view visual information, language instruction and 6D end-effector pose. However, direct retargeting based on end-effector position and orientation trajectories is unsuitable due to workspace discrepancies between Agibot and Unitree G1. For instance, Agibot's arm span reaches 1.8 meters when fully extended, whereas Unitree G1's arm span measures only 1.2 meters. 

To address this, we implement trajectory mapping from Agibot dataset to Unitree official G1 manipulation dataset~\cite{unitreerobotics_G1_ToastedBread_Dataset} , where the latter comprises 2,093 episodes across 7 desktop-level tasks. Specifically, we align the x-axis of the Agibot data with the G1, by applying z-score normalization based on the latter, rescale the y-axis using a scaling factor $\beta=0.6667$ proportional to arm length, and clip the z-axis to [0.15,1.25] with safety bounds.

\subsection{Retargeting in Simulation}
\subsubsection{Model Architecture}

Given that Agibot-World dataset contains extensive household tasks with $z$-coordinates predominantly distributed between 0.2–1.2 m (Fig.~\ref{fig:hot-map}), successful whole-body manipulation necessitates coordinated lower-body motions (e.g., squatting). To address this, we propose a composite hierarchical model for whole-body manipulation retargeting (Fig.~\ref{fig:hierarchical_model}). Specifically:

\begin{figure}[t!]
    \centering
     \includegraphics[width=0.9\linewidth]{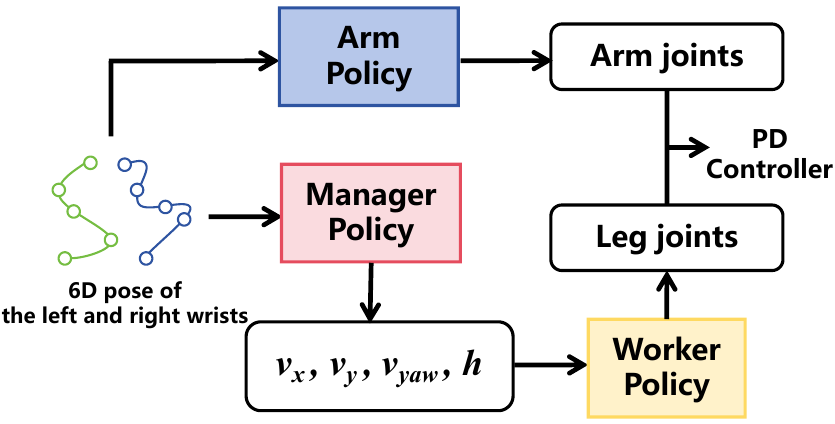}
    \caption{\textbf{Retargeting model architecture.} For whole-body humanoid manipulation with dual end-effector trajectory goals, we decouple control into upper and lower body systems. Arm joints are generated by \textbf{Arm Policy}. For lower-body control, a \textbf{hierarchical model} employs: (1) a \textbf{Manager Policy} that outputs base velocity commands ($v_x$, $v_y$, $v_{\text{yaw}}$) and torso height $h$ based on target wrist poses, and (2) a \textbf{Worker Policy} that executes these commands to control leg joints. }\label{fig:hierarchical_model}
\vspace{-15px}
\end{figure}

\textbf{Arm Policy ($P_\text{arm}$):} Computes target joint angles using Closed-loop Inverse Kinematics (CLIK) via Pinocchio \cite{carpentier2019pinocchio}:
\begin{equation}
    \mathbf{a}^{\text{arm}}_t = P_\text{arm}(\mathbf{T}_{BE}),
    \label{eq_IK}
\end{equation}
with $\mathbf{T}_{BE}$ denoting wrist poses relative to base frame.

% \textbf{Worker Policy ($P_\text{worker}$):} A goal-conditioned RL policy trained following~\cite{ben2025homie} with upper-body motion curriculum for enhanced disturbance robustness.
\textbf{Worker Policy ($P_\text{worker}$):} A goal-conditioned RL policy trained following~\cite{ben2025homie} with an upper-body motion curriculum for enhanced disturbance robustness. Specifically, this curriculum progressively increases the intensity of random upper-body motion disturbances during RL training, thereby improving the robustness of the lower-body locomotion control.
It outputs target joint positions for the 12-DoF lower body:
\begin{equation}
    \mathbf{a}^{\text{leg}}_t = P_\text{worker}(v_x, v_y, v_\text{yaw}, h), 
    \label{eq_worker}
\end{equation}
where $v_x$, $v_y$, $v_\text{yaw}$ control forward/lateral/yaw velocities, and $h$ sets torso height.

\textbf{Manager Policy ($P_\text{manager}$):} Generates lower-body commands from wrist poses:
\begin{equation}
    (v_x, v_y, v_\text{yaw}, h) = P_\text{manager}(\mathbf{T}_{BE}).
    \label{eq_manager}
\end{equation}

The composite hierarchical model $H$ integrates these components:
\begin{equation}
    (\mathbf{a}^{\text{leg}}_t, \mathbf{a}^{\text{arm}}_t) = H(\mathbf{T}_{BE}) = \left( P_\text{worker}\left(P_\text{manager}(\mathbf{T}_{BE})\right), P_\text{arm}(\mathbf{T}_{BE}) \right).
\end{equation}
This model uses the end-effector's pose relative to the robot base, $\mathbf{T}_{BE}$, as its input, and outputs Unitree G1 joint instructions executed via PD controllers.

\begin{algorithm}[t]
\caption{Training Procedure for $P_{\text{manager}}$}
\label{alg:p_manager_training}
\begin{algorithmic}[1]
\Require Seed motion dataset; Mujoco simulator; Isaac Gym simulator
\Ensure Trained $P_{\text{manager}}$ model
\Statex \textbf{Stage 1: Data Preparation}
\State Collect seed trajectories: Initialize Unitree standing, replay upper-limb motions
\State Augment trajectories: Apply PCHIP interpolation to the seed trajectories for height $\in [0.15\,\text{m}, 1.25\,\text{m}]$
\State Generate heuristic commands $\mathbf{a}^*$:
\State \quad $h^* \leftarrow \Delta h_{\text{aug-seed}}$ \Comment{PCHIP-interpolated height}
\State \quad $v_x^* \leftarrow -\Delta p_x,\ v_y^* \leftarrow -\Delta p_y$ \Comment{1s horizon displacement}
\State \quad $v_{\text{yaw}}^* \leftarrow -\Delta \theta$ \Comment{Angular displacement}
\State \quad Clip $v_x \in [-0.8,1.2]$, $v_y \in [-0.5,0.5]$, $v_{\text{yaw}} \in [-1.0,1.0]$
\Statex
\Statex \textbf{Stage 2: Online Learning}
\State Initialize: $\mathcal{D} \gets \emptyset$, $P_m \gets P_{\text{manager}}$, $i \gets 0$
\While{not converged}
\State Execute rollout: $P_m$ runs $T$ steps in $N$ parallel envs
\State Compute $\mathcal{L}_{\text{rollout}} \gets \frac{1}{NT} \sum_{n=1}^N \sum_{t=1}^T \mathcal{L}(P_m(s_t^{(n)}), \mathbf{a_t^{*(n)}})$
\State Update $P_m$ via gradient descent on $\mathcal{L}_{\text{rollout}}$
\If{$i \mod M = 0$} \Comment{Data agg. every $M$ iters}
    \State Aggregate data: $\mathcal{D} \gets \mathcal{D} \cup \{(s_t, \mathbf{a^*_t}) \mid \forall t\}$
    \State Compute $\mathcal{L}_{\text{DA}} \gets \sum_{(s,\mathbf{a^*}) \in \mathcal{D}} \mathcal{L}(P_m(s), \mathbf{a^*})$
    \State Update $P_m$ via gradient descent on $\mathcal{L}_{\text{DA}}$
\EndIf
\State iteration $i \gets i + 1$
\EndWhile
\State \Return $P_{\text{manager}}$
\end{algorithmic}
\end{algorithm}

\subsubsection{Hierarchical Model Training}
The hierarchical model training comprises two stages: $P_{\text{worker}}$ training followed by $P_{\text{manager}}$ training via heuristic-based online learning (Algorithm~\ref{alg:p_manager_training}). Key aspects of $P_{\text{manager}}$ training \lixin{are as follows:}

\textbf{Seed Trajectory Collection:} In MuJoCo, initialize the Unitree G1 standing, replay the upper-limb motion dataset~\cite{unitreerobotics_G1_ToastedBread_Dataset} containing 2,093 episodes, and record the resulting trajectories.

\textbf{Trajectory Augmentation:} Apply PCHIP (Piecewise Cubic Hermite Interpolating Polynomial) interpolation to the seed trajectories to generate height variations $\in [0.15\,\text{m}, 1.25\,\text{m}]$, enabling whole-body manipulation across different heights.

\textbf{Heuristic Target Command ($\mathbf{a^*}$) Generation:} Heuristic ground-truth height targets $h^*$ are derived from the PCHIP-interpolated heights of the seed trajectories. Heuristic velocity commands ($v_x^*$, $v_y^*$, $v_{\text{yaw}}^*$) are computed from the humanoid's base displacement relative to its initial position in Isaac Gym, assuming a 1-second planning horizon.

\textbf{Harmonized Online DAgger:} For brevity, $P_{\text{manager}}$ and the state $s_t$ (representing $T_{\text{BE}}$) are denoted as $P_m$ and $s_t$, respectively. During each iteration, $P_m$ executes a $T$-step rollout ($T=50$) across $N$ parallel environments in Isaac Gym. The loss is minimized as:
        \begin{equation}
        \mathcal{L}_{\text{rollout}} = \frac{1}{N\cdot T} \sum_{n=1}^N \sum_{t=1}^T \mathcal{L}\big( P_m(s_t),\, \mathbf{a^*_t} \big).
        \label{eq:L_rollout}
        \end{equation}
        To mitigate catastrophic forgetting in continual learning, we implement a harmonized Dataset Aggregation (DAgger) strategy. Unlike standard DAgger~\cite{ross2011reduction}, which aggregates data at every iteration, we \textbf{strike a balance} between data efficiency and computational efficiency by subsampling the aggregation process -- specifically, we incorporate new demonstrations only once every   $M=10$ iterations:
        \begin{equation}
        \mathcal{D} \gets \mathcal{D} \cup \left\{ \left( s_t, \mathbf{a^*_t} \right) \mid t \in [1,T] \right\}.
        \label{eq:data_aggregation}
        \end{equation}
        Subsequently, we minimize the aggregated dataset loss:
        \begin{equation}
        \mathcal{L}_{\text{DA}} = \sum_{(s_t,\mathbf{a^*_t}) \in \mathcal{D}} \mathcal{L}(P_m(s_t), \mathbf{a^*_t}).
        \label{eq:L_DA}
        \end{equation}

\begin{figure}[t!]
\centering
 \includegraphics[width=0.95\linewidth]{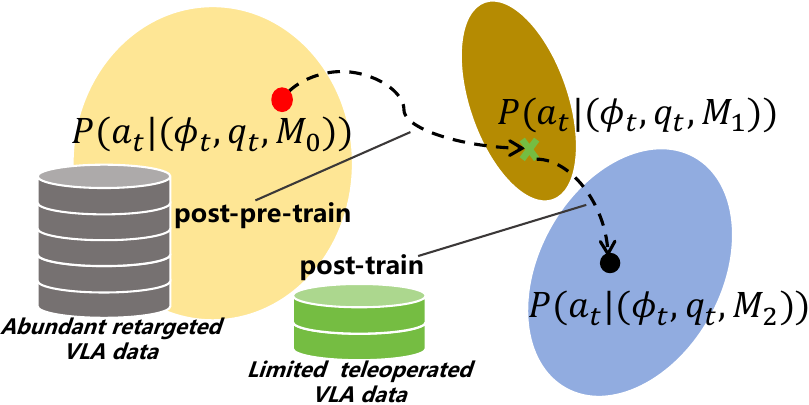}
\caption{\textbf{Two-stage training of the pre-trained VLA.} We first conduct post-pre-training on the pre-trained VLA $M_0$ to obtain $M_1$. This aligns the action distribution $P(a_t \mid (\phi_t, q_t, M_1))$ more closely with the target humanoid's real-world action distribution. Subsequently, we post-train $M_1$ to yield the deployable model $M_2$ for the target humanoid robot.}\label{fig:two_stage_training}
\vspace{-15px}
\end{figure}
% \end{enumerate}
Crucially, this pipeline leverages privileged information unavailable in real deployment. Specifically, in simulation, we can access the torso height corresponding to the current target 6D manipulation trajectory and the humanoid's base displacement relative to the corresponding body position. This privileged information enables efficient generation of heuristic target commands $\mathbf{a^*}$, facilitating effective training of $P_{\text{manager}}$.

\begin{figure*}[t]
\centering
\includegraphics[width=0.99\linewidth]{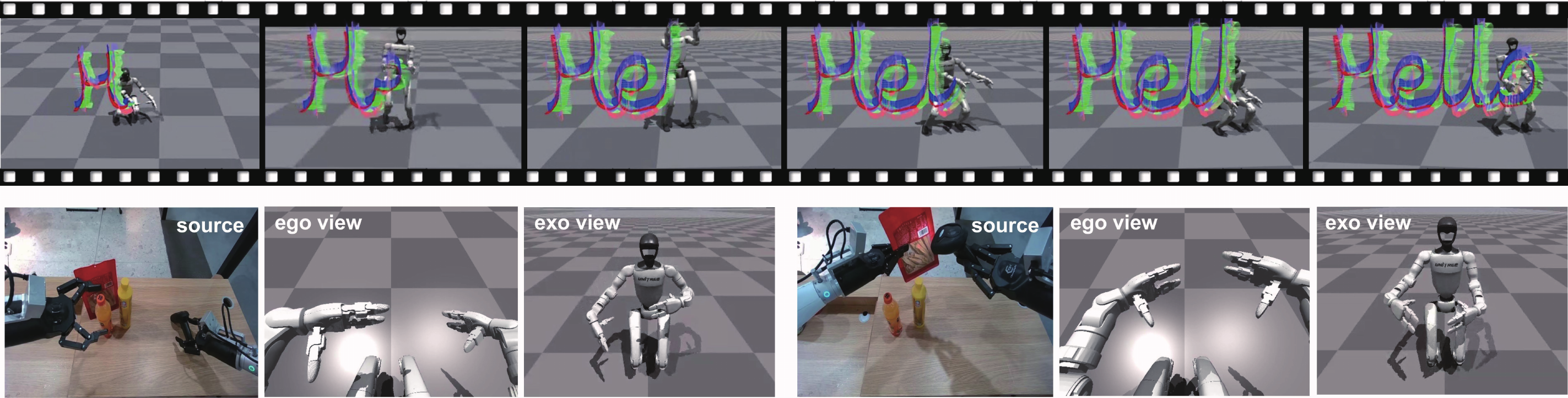}
\caption{\textbf{Real-to-Simulation illustration.} Top: Given target 6D wrist poses (dark trajectories), our retargeting model generates whole-body motion tracking (light trajectories). Bottom: Retargeted Agibot-World manipulation data in Isaac Gym, showing real-world first-person view, simulated first-person view, and simulated third-person view (left to right).}
\label{fig:rea-to-sim}
\vspace{-10px}
\end{figure*}

\subsection{Post-Pre-Training with Retargeted Data}
\label{subsec:post_pre_training}

Post-pre-training (PPT), an intermediate phase between pre-training and post-training, is a widely adopted technique in large language models (LLMs)~\cite{kang2024get, wang2025octothinker} and vision-language models (VLMs)~\cite{yamaguchi2025post}. Similarly, for VLAs, we anticipate that this methodology will also enhance the model’s swift adaptation to downstream tasks, along with enhanced adaptation to and comprehension of the action space, as shown in Fig.~\ref{fig:two_stage_training}.

In this work, we construct multimodal data triplets by integrating retargeted action data with language instructions and visual observations from the original Agibot-World dataset. These triplets are used for post-pre-training of the pre-trained GR00T N1.5 model~\cite{bjorck2025gr00t}.

The post-pre-training phase employs the same objective function as the post-training stage described in~\cite{bjorck2025gr00t}. Given a ground-truth action chunk \( A_{t} \) and sampled noise \( \epsilon \), we construct a noised action chunk:
\(
A_t^{\tau} = \tau A_t + (1 - \tau)\epsilon
\),
where \(\tau \in [0,1]\) denotes the flow-matching timestep. The VLA model \( V_\theta(\phi_t, A_t^{\tau}, q_t) \) predicts the denoising vector field \(\epsilon - A_t\) by minimizing the flow-matching loss:
\begin{equation}
\label{eq:fm_loss}
\mathcal{L}_{\text{fm}}(\theta) = \mathbb{E}_{\tau} \left[ \| V_\theta(\phi_t, A_t^{\tau}, q_t) - (\epsilon - A_t) \|^2 \right].
\end{equation}
Here, \(\phi_t\) represents vision-language token embeddings, \( q_t \) encodes the Unitree G1 whole-body joint state, \(\epsilon\) is Gaussian noise sampled from \(\mathcal{N}(0,I)\), and the expectation is taken over \(\tau\) uniformly distributed in \([0,1]\). 

% During inference, we generate action chunks that span 16 timesteps at 20Hz using 4 denoising steps. Each chunk contains joint position commands for both arms and hands, along with lower-body control commands \((v_x, v_y, v_{\text{yaw}}, h)\) for the \(P_{\text{worker}}\) module, enabling the VLA model to achieve whole-body control of the humanoid. 

% $$
%  v_x, v_y, v_{\text{yaw}}, h, \mathbf{a}^{\text{arm}}_t, \mathbf{a}^{\text{hand}}_t= \text{VLA}(\phi_t, q_t)
%  $$
%  $$
% (\mathbf{a}^{\text{leg}}_t, \mathbf{a}^{\text{arm}}_t, \mathbf{a}^{\text{hand}}_t)  = (P(v_x, v_y, v_{\text{yaw}}, h), \mathbf{a}^{\text{arm}}_t, \mathbf{a}^{\text{hand}}_t)  
% $$

During inference, we generate action chunks that span 16 timesteps at 20Hz using 4 denoising steps. To achieve whole-body control of the humanoid, we adopt a hierarchical control architecture wherein the VLA model predicts high-level commands that are subsequently translated into low-level joint actions by a dedicated worker policy. Specifically, the VLA model outputs base velocity commands $(v_x, v_y, v_{\text{yaw}})$, base height $h$, and joint position targets for the arms and hands:
\begin{equation}
v_x, v_y, v_{\text{yaw}}, h, \mathbf{a}^{\text{arm}}_t, \mathbf{a}^{\text{hand}}_t  =\hat{A_t}= \text{VLA}(\phi_t, q_t).
\end{equation}
These high-level locomotion commands $(v_x, v_y, v_{\text{yaw}}, h)$ are then processed by the worker policy $P_{\text{worker}}$ to generate the corresponding leg joint positions, yielding the complete whole-body action:
\begin{equation}
(\mathbf{a}^{\text{leg}}_t, \mathbf{a}^{\text{arm}}_t, \mathbf{a}^{\text{hand}}_t) = \big(P_{\text{worker}}(v_x, v_y, v_{\text{yaw}}, h), \mathbf{a}^{\text{arm}}_t, \mathbf{a}^{\text{hand}}_t\big).
\end{equation}
This hierarchical decomposition enables the VLA model to focus on high-level task reasoning and end-effector control while delegating the challenging locomotion dynamics to the specialized worker policy.

% During inference, we generate action chunks that span 16 timesteps at 20Hz using 4 denoising steps. To achieve whole-body control of the humanoid, we adopt a hierarchical control architecture wherein the VLA model predicts high-level commands that are subsequently translated into low-level joint actions by a dedicated worker policy. Specifically, the VLA model outputs base velocity commands $(v_x, v_y, v_{\text{yaw}})$, base height $h$, and joint position targets for the arms and hands:
% \begin{equation}
% v_x, v_y, v_{\text{yaw}}, h, \mathbf{a}^{\text{arm}}_t, \mathbf{a}^{\text{hand}}_t = \text{VLA}(\phi_t, q_t).
% \end{equation}

% High-level commands $(v_x, v_y, v_{\text{yaw}}, h)$ are then processed by the worker policy $P_{\text{worker}}$ to generate the leg joint positions. whole-body action as follows：
% \begin{equation}
% (\mathbf{a}^{\text{leg}}_t, \mathbf{a}^{\text{arm}}_t, \mathbf{a}^{\text{hand}}_t) = \big(P_{\text{worker}}(v_x, v_y, v_{\text{yaw}}, h), \mathbf{a}^{\text{arm}}_t, \mathbf{a}^{\text{hand}}_t\big).
% \end{equation}
% This hierarchical decomposition enables the VLA model to focus on high-level task reasoning and end-effector control while delegating the challenging locomotion dynamics to the specialized worker policy.

\subsection{Post-Training}
\label{subsec:post_training}

The post-pre-train stage can be viewed as simulation fine-tuning. However, a visual gap exists between this stage and real-world deployment. Therefore, a final post-training stage, namely real-world fine-tuning, is introduced to enable rapid adaptation of the VLA model.

\textbf{Teleoperation Data Collection. }
We employ the same training methodology as $P_{\text{worker}}$ for lower-body motion generation. However, unlike the hierarchical model $H$, $P_{\text{worker}}$'s control commands originate from human operators via remote control joystick. For upper-body motions (including arm and hand movements), we adopt the Apple Vision Pro-based teleoperation framework proposed in~\cite{cheng2024open} to achieve kinematic mapping. We collect visual data using two wrist RGB cameras (left and right) and one head RGB camera.

\textbf{Fine-Tuning the VLA on the Target Humanoid. }
The collected teleoperation data is utilized to post-train the post-pre-trained VLA model by minimizing the flow-matching loss defined in Eq.~\ref{eq:fm_loss}.

\section{EXPERIMENTS}
Our experimental design addresses four primary questions:

\begin{table}[t!]
\setlength{\abovecaptionskip}{0cm}
\setlength{\belowcaptionskip}{0cm}
\centering
\resizebox{1\linewidth}{!}{%
\begin{tabular}{l cc cc}
\toprule
 \textbf{Whole-body Tracking}  & \multicolumn{2}{c}{\textbf{Mobile}} & \multicolumn{2}{c}{\textbf{Static}} \\
\cmidrule(lr){2-3} \cmidrule(lr){4-5}
\textbf{Methods} & $E_p$ $\downarrow$ & $E_r$ $\downarrow$ & $E_p$ $\downarrow$ & $E_r$ $\downarrow$ \\
\midrule
PPO & 14.326 & 11.853 & 7.964 & 9.160 \\
Standard DAgger & 4.596 & 9.225 & 2.008 & \textbf{5.310} \\
Online learning & 3.764 & 8.085 & 2.073 & 5.365 \\
Online DAgger ($M=1$)  & 3.358 & 7.165 & 2.025 & \textbf{5.327} \\
\textbf{Online DAgger ($M=10$)}  & \textbf{2.851} & \textbf{6.231} & \textbf{1.893} & \textbf{5.331} \\
\bottomrule
\end{tabular}%
}
\caption{\textbf{Whole-body tracking performance comparison.} Lower values ($\downarrow$) indicate better performance for both position ($E_p$) and rotation ($E_r$) errors.}
\label{tab:tracking_performance}
\vspace{-10px}
\end{table}

\textbf{Q1:} How does our hierarchical training framework (with harmonized online DAgger) outperform baselines in humanoid trajectory retargeting? (Section~\ref{subsec:a1})

\textbf{Q2:} Does replacing action data with simulated retargeted actions during fine-tuning \textbf{accelerate real-world adaptation} to humanoid action spaces? (Section~\ref{subsubsec:a2})

\textbf{Q3:} Does post-pre-training enhance \textbf{trajectory generalization} for objects at out-of-distribution positions? (Section~\ref{subsubsec:a3})

\textbf{Q4:} Can post-pre-training unlock \textbf{zero-shot} capabilities for unseen \textbf{manipulation skills} in real-world teleoperation tasks? (Section~\ref{subsubsec:a4})

\subsection{Evaluation on retargeting model}
\label{subsec:a1}

\begin{table*}[t!]
\setlength{\abovecaptionskip}{0cm}
\setlength{\belowcaptionskip}{0cm}
\centering
% \resizebox{0.85\linewidth}{!}{%
\small
\begin{tabular}{l c c c c c}
\toprule
\textbf{Tasks} & \textbf{Height (cm)} & \textbf{Teleop SR} & \textbf{w. PPT SR} & \textbf{w/o. PPT SR} & \textbf{w/o. PPT SR} \\
& & & \textbf{(3K steps)} & \textbf{(3K steps)} & \textbf{(10K steps)} \\
\midrule
Pick Mickey Mouse & 39 & 5/7 (71\%) & \textbf{100\%} & 0\% & 80\% \\
Store Toys & 55 & 10/13 (77\%) & \textbf{70\%} & 0\% & 30\% (+30\%) \\
Clean the Table & 55 & 6/11 (55\%) & \textbf{70\%} & 0\% & \textbf{70\%} \\
Pick Orange \& Place & 76 & 7/11 (64\%) & \textbf{10\% (+ 50\%)} & 0\% & 0\% (+ 30\%) \\
\bottomrule
\end{tabular}%
% }
\caption{\textbf{Success rates for four teleoperated tasks.} Height indicates the elevation of the operation plane relative to the robot's standing surface. Teleop SR includes both the number of demonstrations collected and the operator's teleoperation success rate. For \emph{Store Toys}, parenthetical rates indicate partial success (toys placed on box edges). For \emph{Pick Orange \& Place}, they denote successful grasping but failed placement. The lightweight orange toy induced vibrations during autonomous operation, resulting in reduced task success rates.}
\label{tab:in_domain_4tasks_performance}
\vspace{-10px}
\end{table*}

\textbf{Baselines. }
The hierarchical model was trained using Harmonized Online DAgger. To validate this approach's efficacy and efficiency for tracking model training, we compared against several baselines: reward-based PPO, standard DAgger ($M=1$  with only Eq.~\ref{eq:L_DA}), online learning (only Eq.~\ref{eq:L_rollout}), and Standard Online DAgger ($M=1$  with both Eq.~\ref{eq:L_rollout} and Eq.~\ref{eq:L_DA}).
% standard DAgger($M=1$ with only ~\ref{eq:L_rollout}), online learning (only ~\ref{eq:data_aggregation}), standard online DAgger ($M=1$ with ~\ref{eq:data_aggregation} and~\ref{eq:L_rollout}).

\textbf{Implementation Details. }
All experiments employed 512 parallel environments with 200 training iterations, except PPO which used 800 iterations to account for its additional value model training. Training and inference were conducted on a single RTX 4090 GPU with Intel Core i9-14900K CPU.

\textbf{Superior Tracking Performance. }
% \label{subsubsec:a1}
Tracking performance was evaluated through simulation by computing Mean Absolute Error (MAE) in position $E_p$ (cm) and rotation $E_r$ (degrees) between the Unitree G1's wrist trajectory and target trajectory. 
We include PPO as a baseline to demonstrate that even with suboptimal heuristic-based ground truth, imitation learning achieves faster convergence and more stable training compared to reinforcement learning. While PPO theoretically could reach higher asymptotic performance with sufficient exploration, we empirically observe slow convergence and inferior sample efficiency in practice, likely due to reward sparsity in loco-manipulation tasks. 
As shown in Table~\ref{tab:tracking_performance}, our harmonized online DAgger ($M=10$) achieves lower storage usage and higher learning efficiency while enabling loco-manipulation tracking capabilities (Fig.~\ref{fig:rea-to-sim}).

\subsection{Evaluation on VLA with post-pre-training}

\begin{figure}[t!]
    \centering
     \includegraphics[width=0.99\linewidth]{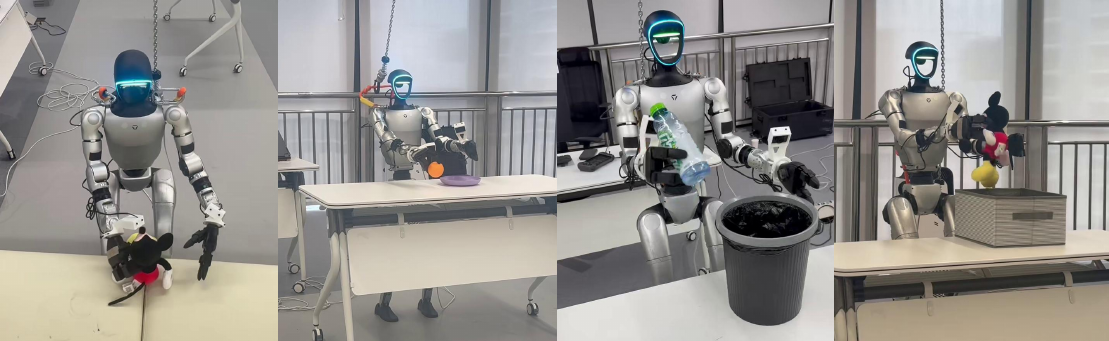}
    \caption{\textbf{Overview of four teleoperated tasks.} The experiments comprise pick-and-place operations at varying heights, requiring continuous height adjustments.}\label{fig:four-tasks}
\vspace{-15px}
\end{figure}

\textbf{Datasets. } 
For each task within the Agibot-World beta dataset, 10 episodes were randomly sampled. Tasks involving both dexterous hands and parallel grippers underwent separate sampling, with 10 episodes selected per end-effector type. All valid episodes were included for tasks containing fewer than 10 episodes, while episodes exhibiting frame errors were systematically excluded. This procedure yielded a dataset comprising 176 distinct tasks and 1960 episodes, corresponding to approximately 35 hours of simulated interaction. We then preprocessed these data using the method described in \ref{subsec:data_preprocess}.
Leveraging Isaac Gym's environmental parallelism, we performed hierarchical body motion retargeting across all episodes within tens of minutes. Building upon these retargeted body motions, we implemented end-effector mapping: Agibot-World uses grippers (85\% trajectories) and hands (15\%), so we map thumb/index finger opening to gripper space; for target end-effector Unitree Dex-3 (7-DOF hand), pre-collected open/close joint positions serve as retargeting targets.The retargeted Unitree G1 motions replaced original actuator commands, generating multimodal (action, language, vision) data triplets.
For subsequent post-training, 28 episodes of real-world whole-body manipulation data were collected using a Unitree G1 humanoid robot across four distinct height configurations, as shown in Fig.~\ref{fig:four-tasks}. This dataset comprises approximately 10 minutes of operational time.

\textbf{Baselines. } We establish VLAs without post-pre-training as baselines: models directly post-trained on the pre-trained GR00T N1.5 for 3K and 10K steps. These are compared against our post-pre-trained VLA post-trained for 3K steps.
 
\textbf{Implementation Details. }
Post-pre-training used retargeted action-vision-language triplets on dual A100 80GB GPUs (batch size=128, 60K steps); Post-training employed real-world whole-body manipulation data on a single A100 GPU (batch size=16, 3K steps).
Concurrently, two control models were trained from the GR00T N1.5 checkpoint (using only real-world data): a 3K-step variant and a 10K-step variant, both trained on a single A100 GPU with a batch size of 16.

\subsubsection{Accelerated Adaptation to Humanoid Action Space}
\label{subsubsec:a2}
Unlike LLMs and VLMs, post-pre-training the VLA model in this work involves addressing the sim-to-real gap and visual inconsistencies. To evaluate whether our post-pre-training methodology genuinely facilitates accelerated downstream post-training adaptation, akin to its application in VLMs and LLMs, we evaluated these models on four real-world tasks, with 10 experimental trials conducted per task. Results (Table~\ref{tab:in_domain_4tasks_performance}) indicate that when target objects were positioned \textit{in-domain} (matching the locations in the real-world dataset), the model subjected to post-pre-training followed by only 3K steps of post-training achieved a higher success rate than the model trained solely on real-world data for 10K steps. Notably, the model trained exclusively on real-world data for 3K steps failed to learn the task effectively; during execution, it exhibited only oscillatory behavior near the target without demonstrating any intent to grasp the object.

\subsubsection{Enhanced Trajectory Generalization}
\label{subsubsec:a3}
While section~\ref{subsubsec:a2} demonstrated that post-pre-training accelerates convergence during downstream post-training, we further investigated whether it enhances the model's understanding of the target humanoid action space, thereby improving trajectory generalization. Task difficulty was increased by relocating the target object. We placed the Mickey Mouse in the \textit{Pick Mickey Mouse} task at positions unseen during teleoperation and conducted 10 trials, with success rates under these novel configurations reported in Table~\ref{tab:out_of_domain}.

\begin{table}[t!]
\setlength{\abovecaptionskip}{0cm}
\setlength{\belowcaptionskip}{0cm}
\centering
\resizebox{0.99\linewidth}{!}{%
\begin{tabular}{l c c}
\toprule
\textbf{Models} & \textbf{Success Rate} $\uparrow$  & \textbf{DTW Distance} $\uparrow$  \\
\midrule
w. PPT (3K steps) & \textbf{80\% }& \textbf{0.278} \\
w/o. PPT (10K steps) & 0\% & 0.220 \\
\bottomrule
\end{tabular}%
}
\caption{\textbf{Impact of PPT on unseen object placements.}}
\label{tab:out_of_domain}
\vspace{-10px}
\end{table}

\begin{figure}[t!]
\centering
     \includegraphics[width=0.99\linewidth]{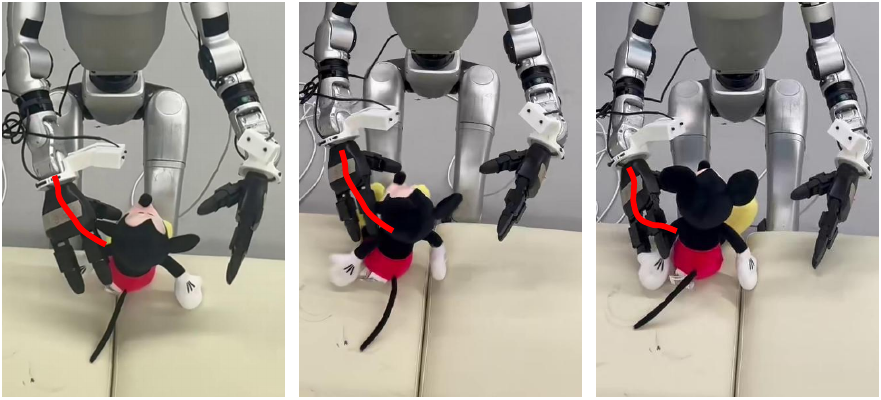}
    \caption{\textbf{Qualitative analysis of trajectory generalization.} 
    The left image shows Mickey Mouse positioned consistently with the teleoperation data, while the middle and right images show placement closer to the humanoid's right hand. Trajectory analysis reveals that (middle) without PPT, the VLA mimics teleoperated motions (left), approaching from above, whereas (right) with PPT, the VLA adapts to grasp from below.}
    \label{fig:Qualitative_analysis}
\vspace{-15px}
\end{figure}

The observed decline in success rate for the model trained without post-pre-training suggests a propensity for overfitting to the specific trajectories encountered during post-training. To substantiate this hypothesis, the joint trajectories of the right arm generated during \textit{Pick Mickey Mouse} execution were recorded for both the post-pre-trained + 3K steps post-trained model and the model trained solely on real-world data for 10K steps. The FastDTW algorithm~\cite{salvador2007toward} was employed to compute the similarity between these generated trajectories and the corresponding real-world teleoperation data. The model trained without post-pre-training exhibited a smaller DTW distance as shown in Table~\ref{tab:out_of_domain}, indicating closer replication of the memorized teleoperation trajectories. This memorization bias explains its reduced robustness to spatial variations. Qualitative analysis of the Unitree G1's execution (Fig.~\ref{fig:Qualitative_analysis}) further corroborates this interpretation.

\subsubsection{Unlocking Zero-shot Skill Generalization}
\label{subsubsec:a4}

\begin{figure}[t!]
    \centering
     \includegraphics[width=0.99\linewidth]{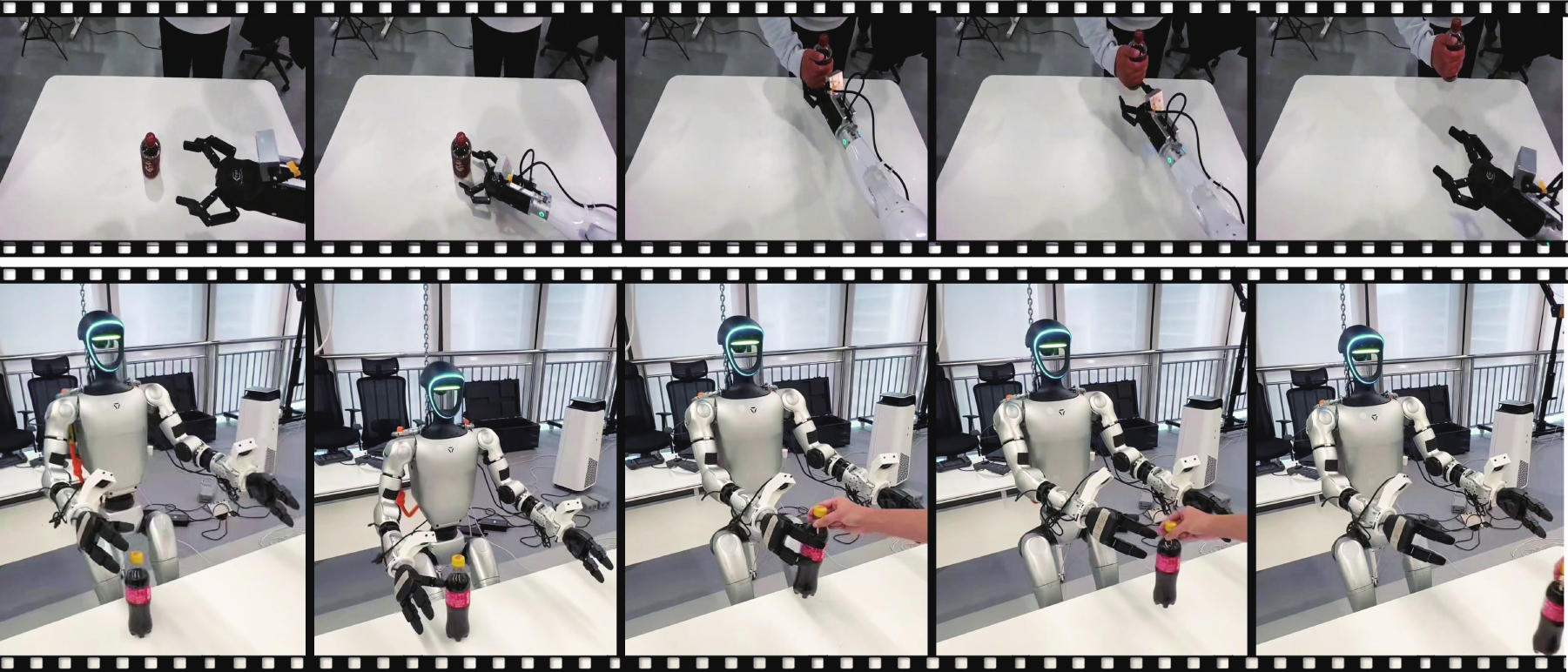}
     \caption{\textbf{Zero-Shot execution of \textit{Pass the Water} task.} Top: First-person view examples of the \textit{Pass the Water} task from the Agibot-World dataset. Bottom: The post-pre-trained VLA successfully executes this task despite receiving no fine-tuning using real-world teleoperation data for this specific task.}\label{fig:zero-shot} 
\vspace{-15px}
\end{figure}

Beyond enhancing action space comprehension, we investigated whether the post-pre-training phase improves task generalization, enabling the execution of skills absent from the teleoperation dataset. The model was evaluated on the task \textit{Pass the Water}, which was included during post-pre-training but excluded from the real-world teleoperation dataset. Remarkably, as shown in Fig.~\ref{fig:zero-shot}, the model successfully executed this task on the Unitree G1 robot in a \textbf{zero-shot} manner, demonstrating acquired generalization capabilities.

\section{CONCLUSIONS AND LIMITATIONS}
This paper tackles the challenge of data scarcity in training vision-language-action (VLA) models for whole-body manipulation with bipedal humanoids. We introduce TrajBooster, an end-effector trajectory–driven real-to-sim-to-real pipeline designed to enhance cross-embodiment VLA performance. Leveraging harmonized online DAgger, our approach uses privileged simulator information together with heuristic methods to efficiently train a hierarchical model that retargets large-scale, heterogeneous robot data10 minutes of teleoperation data for this post-training stage, TrajBooster enables rapid adaptation to new action spaces, robust trajectory generalization, and zero-shot transfer to previously unseen scenarios.

As an early investigation into VLA for bipedal humanoid whole-body manipulation, the following limitations highlight key directions for future improvement:

\begin{enumerate}
    \item \textbf{End-effector limitation.} 
    The Unitree Dex-3 restricts tasks to simple pick-and-place due to limited precision. Future work will employ dexterous hands with tactile sensing for advanced manipulation.

    \item \textbf{Action-visual consistency.} 
    Our method only replaces the action space while retaining visual input. We will explore embodiment alignment in visual observations to improve perception-action consistency in the future.

    \item \textbf{Loco-manipulation data scarcity.} 
    The lack of large-scale loco-manipulation data confines our study to mostly stationary tasks. Future work will extend the framework to richer mobile scenarios.

    \item \textbf{Scaling limitations.} 
    The experiments in this work are limited by the scale of the dataset and computational resources. We intend to incorporate more heterogeneous data in the future, going beyond the Agibot G1 robot and Agibot-World dataset for retargeting.
\end{enumerate}

\section*{ACKNOWLEDGMENT}

This work was supported by the Brain Science and Brain-like Intelligence Technology — National Science and Technology Major Project (Grant No. 2022ZD0208800), National Natural Science Foundation of China (Grant No. 62506232), STCSM “Yangfan” Program (Grant No. 24YF2722000), Science and Technology Major Project of Jiangsu Province (Grant No. BG2024041). The authors also gratefully acknowledge Dr. Yue Gao for generously providing access to the robotic platform and experimental facilities that made this work possible.

\bibliographystyle{IEEEtran}
\bibliography{references}

\end{document}